\documentclass[letterpaper, conference]{ieeeconf}

\IEEEoverridecommandlockouts
\usepackage{graphicx}
\usepackage{picinpar}
\usepackage{amsmath,bm}
\usepackage{url}
\usepackage{colortbl}
\usepackage{soul}
\usepackage{multirow}
\usepackage{pifont}
\usepackage{color}
\usepackage{alltt}
\usepackage[hidelinks]{hyperref}
\usepackage{enumerate}
\usepackage{breakurl}
\usepackage{epstopdf}
\usepackage{pbox}
\usepackage{indentfirst}
\usepackage{mathtools}
\usepackage{booktabs}
\usepackage{makecell}
\usepackage{subfigure}
\usepackage{float}
\usepackage{threeparttable}
\usepackage{algorithmic}
\usepackage{algorithm}
\usepackage{amsfonts}
\usepackage[allow-number-unit-breaks]{siunitx}
\usepackage[dvipsnames]{xcolor}
\usepackage{cite}

\newcommand{\etalcite}[1]{et al.~\cite{#1}}

\title{\LARGE \bf DogLegs: Robust Proprioceptive State Estimation for Legged Robots Using Multiple Leg-Mounted IMUs}

\author{Yibin~Wu$^{1, 2}$ \quad ~Jian~Kuang$^{4}$ \quad ~Shahram~Khorshidi$^{1, 3}$ \quad ~Xiaoji~Niu$^{4}$ \\ ~Lasse~Klingbeil$^{1, 2}$ \quad ~Maren~Bennewitz$^{1, 3}$ \quad ~Heiner~Kuhlmann$^{1, 2}$
\thanks{$^{1}$ Center for Robotics, University of Bonn, Germany}%
\thanks{$^{2}$ Institute of Geodesy and Geoinformation, University of Bonn, Germany}%
\thanks{$^{3}$ Humanoid Robots Lab, University of Bonn, Germany}%
\thanks{$^{4}$ GNSS Research Center, Wuhan University, China}%
\thanks{Corresponding: {\tt\small yibin.wu@igg.uni-bonn.de}}
}

\begin{document}
\maketitle

\begin{abstract}
Robust and accurate proprioceptive state estimation of the main body is crucial for legged robots to execute tasks in extreme environments where exteroceptive sensors, such as LiDARs and cameras, may become unreliable. In this paper, we propose DogLegs, a state estimation system for legged robots that fuses the measurements from a body-mounted inertial measurement unit (Body-IMU), joint encoders, and multiple leg-mounted IMUs (Leg-IMU) using an extended Kalman filter (EKF). The filter system contains the error states of all IMU frames. The Leg-IMUs are used to detect foot contact, thereby providing zero-velocity measurements to update the state of the Leg-IMU frames. Additionally, we compute the relative position constraints between the Body-IMU and Leg-IMUs by the leg kinematics and use them to update the main body state and reduce the error drift of the individual IMU frames. Field experimental results have shown that our proposed DogLegs system achieves better state estimation accuracy compared to the traditional leg odometry method (using only Body-IMU and joint encoders) across various terrains. We make our code and datasets publicly available to benefit the research community (https://github.com/YibinWu/DogLegs).
\end{abstract}

\section{Introduction}

Legged robots rely on accurate state estimation to keep balance and navigate challenging environments~\cite{bloesch2013ekf, yangshuo2023mipo, camurri2017ral, lin2021corl}. To this end, proprioceptive sensors, such as inertial measurement units (IMUs) and joint encoders, play an essential role. On one hand, they provide high-frequency egomotion information that can be fused with the exteroceptive sensors, such as LiDARs and cameras~\cite{wisth2023tro,shuoyang2023cerberus,ou2024legkilo}, to enhance localization accuracy. On the other hand, they serve as a backup solution when the exteroceptive sensors fail, such as in dark environments or under canopy conditions in agricultural applications.  

Current approaches to this problem primarily use the robot body-mounted IMU (Body-IMU), joint encoders and foot force sensors~\cite{bloesch2013ekf, bloesch2013ukf, lin2021corl,hartley2020ijrr}. In these systems, the Body-IMU initially propagates the robot's main body state, which is then refined by integrating leg kinematics estimated from joint encoders upon detecting foot contact using the foot force sensor. The most widely used approach for foot contact detection relies on empirically thresholding force sensor readings, but this method is susceptible to errors due to slippage and sensor degradation over time. In addition, the foot positions during swing phase cannot be updated~\cite{bloesch2013ekf, yangshuo2023mipo} and need to be relocated when the foot regains contact with the ground, introducing further error into the robot state estimation.


Recent studies have attempted to mount IMUs on robot legs (Leg-IMUs) to detect foot contact and improve state estimation. Yang~\etalcite{yangshuo2023mipo} proposed to mount multiple IMUs on the robot feet to provide additional foot velocity measurements within the similar extended Kalman filter (EKF) structure as in Bloesch~\etalcite{bloesch2013ekf,bloesch2013ukf}. However, the state estimation capability of the Leg-IMUs had not been fully utilized because IMU can provide six degrees of freedom pose estimation, not only velocity measurement.

\begin{figure}[t]
    \setlength{\abovecaptionskip}{0pt} 
    \setlength{\belowcaptionskip}{0pt} 
	\centering
	\includegraphics[width=8.8cm]{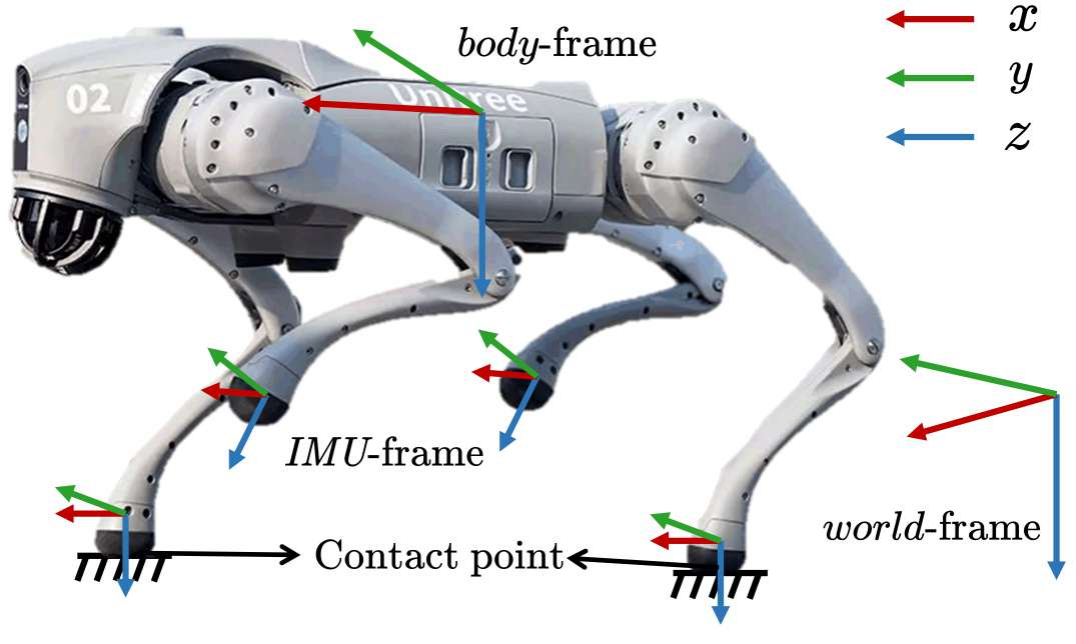}
    \caption{Illustration of the installation of the multiple Leg-IMUs and the coordinate systems of our proposed DogLegs on a quadruped robot. Our approach fuses the measurements from a Body-IMU, multiple Leg-IMUs and joint encoders. The Leg-IMUs are mounted on the lower legs of the robot and close to the feet. The robot \textit{body}-frame is defined with the origin at the center of the robot body and aligns with the coordinates of the Body-IMU. The \textit{world}-frame is defined with the \textit{body}-frame at the initial position.}
	\label{fig:coordinates}
\end{figure}

The main contribution of this paper is the development of DogLegs, a framework that fuses the full state estimation information from all the individual IMU systems, including the Body-IMU and multiple Leg-IMUs, through an EKF. Specifically, each IMU system independently performs state propagation while a comprehensive error-state vector is maintained to account for errors across all IMU frames. The Leg-IMUs are used for foot contact detection, providing zero-velocity measurements to update the state of the Leg-IMU frames. Additionally, we use leg kinematics to compute the relative position constraints between the Body-IMU and Leg-IMUs. This measurement helps update the main body state, reduce error drift in individual IMU frames, and estimate and compensate for IMU bias. Fig.~\ref{fig:coordinates} illustrates the installation of multiple Leg-IMUs and the coordinate systems involved within our proposed DogLegs framework, using a quadruped robot as an example. 

In summary, we make two major claims:
\begin{itemize}
    \item Doglegs achieves better performance in state estimation compared to the traditional EKF-based leg odometry (using only a Body-IMU and joint encoders) across different terrains.
    \item DogLegs reduces the error drift of the individual IMU systems by incorporating the relative position constraints between the multiple IMUs.
\end{itemize}
Moreover, we make our code and datasets publicly available to benefit the research community.

\section{Related Work}

\subsection{Proprioceptive Legged Robot State Estimation}
State estimation for the main body of legged robots has received significant attention in recent years. While numerous sensor fusion-based systems leveraging various exteroceptive sensors, e.g., cameras~\cite{shuoyang2023cerberus, wisth2019ral, kim2022ral} and LiDARs~\cite{ou2024legkilo,wisth2023tro,camurri2020pronto}, have been proposed, this paper focuses specifically on proprioceptive sensors, including IMUs and joint encoders.

An early approach to proprioceptive state estimation for legged robots was proposed by Bloesch~\etalcite{bloesch2013ekf}, who employed an EKF to fuse outputs from a Body-IMU, joint encoders, and foot force sensors. This method augmented the foot contact position into the state and updated it with the main body state using the leg kinematics when the foot contacts ground. This approach has become the basis for many subsequent state estimation frameworks. Later, Bloesch~\etalcite{bloesch2013ukf} extended this work by incorporating the zero-velocity measurement and replacing the EKF with an unscented Kalman filter to improve accuracy and robustness. 

Taking advantages of the symmetry of state estimation problem, researchers have introduced the invariant extended Kalman filter (InEKF) into the legged robot state estimation, using Lie Groups and Lie algebra for state propagation and error tracking~\cite{hartley2018rss, hartley2020ijrr, lin2021corl, teng2021icra, kim2021ral, yoon2024tro}. Hartley~\etalcite{hartley2018rss, hartley2020ijrr} derived a continuous time right-invariant EKF for an IMU/contact process model with a forward kinematic measurement model. This method used the same sensor input as Bloesch~\etalcite{bloesch2013ekf,bloesch2013ukf} but with a different state estimator. Lin~\etalcite{lin2021corl} trained a network to estimate the foot contact and used it in a similar InEKF framework. Kim~\etalcite{kim2021ral} and Yoon~\etalcite{yoon2024tro} extended InEKF into a fixed-lag invariant smoother, coupled with a foot slippage detection method, which improved state estimation accuracy by optimizing robot states within a sliding window. 

To improve the sensing ability of legged robots, additional IMUs have been incorporated into their lower legs. Kolvenbach~\etalcite{kolvenbach2020jfr} embedded IMUs in robot feet to inspect concrete deterioration through a specialized scratching motion; however, these IMUs were not utilized for main body state estimation. Maravgakis~\etalcite{Michael2023contact} proposed estimating the probability of stable foot contact with Leg-IMUs by employing a kernel density estimator to approximate the probability density function. Yang~\etalcite{yangshuo2023mipo} placed multiple IMUs on the robot feet to provide additional foot velocity measurements within the similar EKF structure as Bloesch~\etalcite{bloesch2013ekf}. However, the potential of IMUs for full attitude estimation was overlooked, and the sensor bias of the multiple IMUs was also not addressed in this study. 

In summary, existing legged robots equipped with Leg-IMUs have yet to fully leverage the full-state pose estimation capabilities of the Leg-IMUs. This leaves significant potential for integrating multiple IMUs to reduce random errors of the IMUs, thereby enhancing the accuracy and robustness of main body state estimation.
\begin{figure*}[t]
	\centering
	\includegraphics[width=18cm]{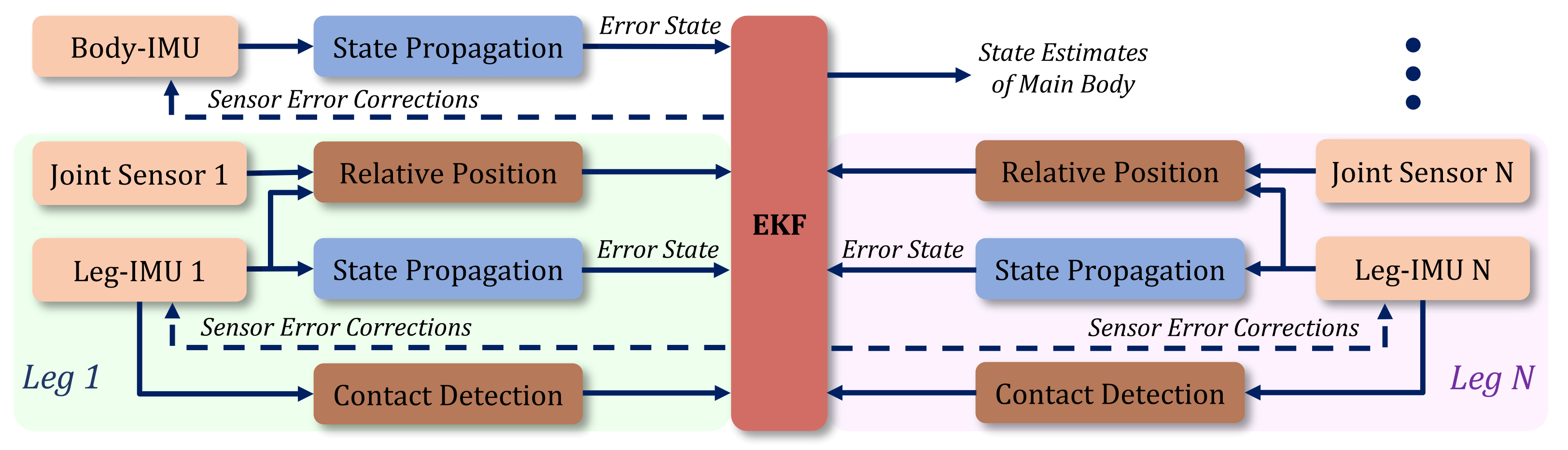}
    \caption{Overview of our proposed DogLegs system. We fuse the measurements from a Body-IMU, joint encoders, and multiple Leg-IMUs to estimate the state of the main body for legged robots via an error-state EKF. The system model of our EKF framework contains the error states from all IMUs. $N$ represents the number of legs. Joint encoders on each leg provide the relative position constraints between the Body-IMU and the corresponding Leg-IMU. Additionally, we refine the IMU outputs by compensating for estimated sensor errors.}
	\label{fig:overview}
    \vspace{-0.1cm}
\end{figure*}
\subsection{Foot-Mounted IMU-based Pedestrian Navigation}

Our approach of utilizing Leg-IMUs for foot contact detection and subsequent state updates is inspired by advancements in foot-mounted inertial navigation systems (Foot-INS)~\cite{foxlin2005,niu2022iot,johan2021sj,kuang2025ieeesj} for pedestrian localization. By detecting periodic foot-ground contact using a foot-mounted IMU, Foot-INS exploits zero-velocity measurements to mitigate IMU error drift. This method is primarily used in professional indoor applications, such as tracking first responders and tunnel inspectors. 

Despite its widespread application in pedestrian indoor navigation and its effectiveness in reducing IMU drift, Foot-INS has seen limited application in legged robots. Unlike humans, legged robots, especially quadrupeds exhibit distinct gaits with less stable foot-ground contact. Moreover, they often operate in challenging terrains, introducing estimation errors when applying conventional pedestrian Foot-INS methods. Therefore, adapting Foot-INS to legged robots remains an open research challenge.

To address this, we propose DogLegs, a proprioceptive state estimation system that integrates diverse motion data from multiple IMUs, including a Body-IMU and Leg-IMUs mounted on each leg of the robot. Our approach demonstrates reduced drift compared to traditional EKF-based leg odometry and Foot-INS across different terrains. 

\section{Our Approach}
Legged robots such as quadrupeds typically have spherical point feet. Due to this design, the foot and the corresponding lower leg of a quadruped robot can be considered as a single rigid body. Therefore, we mount the IMUs on the lower legs, close to the feet (as shown in Fig.~\ref{fig:experimental_robot}) and refer to them as Leg-IMUs throughout this paper. For robots with ankle joints, the IMUs should be mounted on the feet to maintain direct ground contact.

Fig.~\ref{fig:overview} illustrates the algorithmic structure of our proposed DogLegs system. Our method is implemented using an error-state EKF to fuse the measurements from joint encoders and all the IMUs. The Leg-IMUs propagate the states of their mounting points on the legs, while the Body-IMU propagates the state of the main body. Simultaneously, the Leg-IMUs detect foot contact, i.e., zero-velocity, to update their respective states. Additionally, relative position constraints between the Body-IMU and Leg-IMUs (estimated with joint encoders) are incorporated to mitigate error accumulation across all IMUs. We use the estimated IMU sensor errors to correct the IMU readings.

\subsection{Foot Contact Detection}
We detect the robot foot contact by analyzing the Leg-IMU readings~\cite{johan2021sj} rather than force sensor measurements. The main reason is that IMUs provide richer motion information, capturing six degrees of freedom, whereas foot force sensors are typically limited to measuring only the normal contact force in many robots, e.g., Unitree Go1 and Go2 robot\footnote{\url{https://shop.unitree.com/}}. 

Given the fact that the foot acceleration equals to zero at the moment of contact, we employ the generalized likelihood ratio test \cite{kuang2025ieeesj, isaac2010zupt,niu2022iot} using a batch of accelerometer measurements as the foot contact detection criterion, which is given by

\begin{equation}
\frac{1}{2M+1} \sum_{i=k-M}^{k+M} \left\| \bm{f}_{i} - \bm{g} \frac{\bar{\bm{f}}}{\|\bar{\bm{f}}\|} \right\| \leq \bm{\gamma},
\end{equation}
where, $2M+1$ is the batch size; $\bm{f}_{i}$ is the accelerometer measurement at time $i$; $\bm{g}$ is the gravity vector; $\bar{\bm{f}}$ is the average accelerometer measurement within the batch; $\|\cdot\|$ is the L2 norm of a vector; $\bm{\gamma}$ is an empirically determined threshold.
\subsection{Error State Model}
Given gyroscope and acceleration measurements at each time step, the state of the rigid platform on which the IMU is mounted can be propagated by integrating the IMU data. In this paper, we employ the strapdown inertial navigation system (INS) \cite{Shin2005, Groves2013,wu2024icra} for each IMU to propagate its own state. Meanwhile, to mitigate system nonlinearity, we adopt the error-state representation~\cite{niu2021} as the system model within the EKF framework. The whole error-state vector maintained in the EKF is given by
\begin{equation}
    \bm{\mathcal{X}} = \left[\delta \bm{x}_0^{\top} \ , \ \delta \bm{x}_1^{\top} \ , \ \cdots \ , \ \delta \bm{x}_{N}^{\top} \right]^{\top},
\end{equation}
where $N$ represents the number of legs, and ${\delta \bm{x}}$ represents the error state of each IMU, which is defined as
\begin{equation}
    {\delta \bm{x}}=\left[\delta \bm{p}^{\top} \ , \ \delta \bm{v}^{\top} \ , \ \delta \bm{\phi}^{\top} \ , \ \bm{b}_{g}^{\top} \ , \ \bm{b}_{a}^{\top} \right]^{\top} \in \mathbb{R}^{15},
    \label{statevector}
\end{equation}
where $\delta \bm{p}$, $\delta \bm{v}$ and $\delta \bm{\phi}$ denote the position, velocity, and attitude errors respectively, and $\bm{b}_{g}$ and $\bm{b}_{a}$ denote the residual bias errors of the accelerometer and gyroscope, respectively. We integrate the error states from each IMU into a single vector, which serves as the system model in our EKF pipeline~\cite{wu2022tits}. Consequently, the total dimension of the error state vector is $15~\!(N+1)$. The continuous-time error state propagation model used in this paper is given by
\begin{equation}
    \left\{
    \begin{aligned}
    \delta \dot{\bm{p}}&=\delta \bm{v} \\
    \delta \dot{\bm{v}}&=\mathbf{R}\bm{f} \times \delta\bm{\phi} + \mathbf{R}{\bm{b}}_{a} \\
    \delta\dot{\bm{\phi}}&=-\mathbf{R}{\bm{b}}_{g}\\
    \dot{\bm{b}}_{g}&=-(1/\tau_{bg}) \bm{b}_{g}+\bm{n}_{bg}\\
    \dot{\bm{b}}_{a}&=-(1/\tau_{ba}) \bm{b}_{a}+\bm{n}_{ba}\\
    \end{aligned},
    \right.
    \end{equation}
where $\mathbf{R}$ is the rotation matrix of the IMU with respect to (w.r.t.) the reference frame; $\bm{f}$ is the accelerometer measurement in the \textit{IMU}-frame; $\tau_{bg}$ and $\tau_{ba}$ are the correlation time in the first-order Gauss-Markov model~\cite{Shin2005} of the gyroscope bias and accelerometer bias, respectively; $\bm{n}_{bg}$ and $\bm{n}_{ba}$ denote the white noise of the gyroscope bias and accelerometer bias, respectively. This error state propagation model is applied to all the IMUs.

\subsection{Observation Model}
\subsubsection{Zero-Velocity Update}
Generally, when the foot is in contact with the ground without slippage, its velocity in the \textit{world}-frame should be zero. This condition can be formulated as a measurement model~\cite{wu2021tvt} and incorporated into the Leg-IMU state estimation. However, it is important to note that only the velocity of the contact point on the foot is strictly zero, while other points on the foot may still be in motion. Since the Leg-IMU measures velocity at its mounting location rather than at the contact point, we must transform the velocity estimated by the Leg-IMU to the contact point. 

The velocity of the contact point on the robot foot in the \textit{world}-frame indicated by the Leg-IMU can be expressed as
\begin{equation}
    \hat{\bm{v}}^{w}_{f} = \hat{\bm{v}}^{w}_\text{limu} + \hat{\mathbf{R}}^{w}_\text{limu}\left(\hat{\bm{\omega}}_\text{limu}\times\right)\hat{\bm{l}}_{f}^\text{limu},
\end{equation}
where $\hat{\bm{v}}^{w}_\text{limu}$ is the velocity estimates of the Leg-IMU in the \textit{world}-frame; $\hat{{\mathbf{R}}}^{w}_\text{limu}$ is the rotation matrix of the Leg-IMU; $\hat{\bm{\omega}}_\text{limu}\times$ is the skewsymmetric matrix of the angular velocity measurement of the Leg-IMU; $\hat{\bm{l}}_{f}^\text{limu}$ is the position of the foot contact point in the Leg-IMU frame, as shown in Fig.~\ref{fig:coordinates}. This position misalignment between the Leg-IMU and the foot contact point can be measured in advance.

When the foot contacts the ground, the observed velocity of the contact point in the \textit{world}-frame is
\begin{equation}
    \tilde{\bm{v}}^{w}_{f} = \mathbf{0} - \bm{n}_{v^w_{f}},
\end{equation}
where $\bm{n}_{v^w_{f}}$ indicates the noise. Consequently, we can build the zero-velocity measurement model as 
\begin{equation}
    \begin{aligned}
        \delta\bm{z}_v &= \hat{\bm{v}}^{w}_f - \tilde{\bm{v}}^{w}_f\\
        &= \delta\bm{v}^{w}_\text{limu} + \mathbf{R}^{w}_\text{limu}\left(\hat{\bm{\omega}}_\text{limu}\times\right)\delta\bm{\phi}_\text{limu} - \mathbf{R}^{w}_\text{limu}\left({\bm{l}}_{f}^\text{limu}\times\right){\bm{b}}_{g},
    \end{aligned}
\end{equation}
where $\delta\bm{v}^{w}_\text{limu}$, $\delta\bm{\phi}_\text{limu}$, and ${\bm{b}}_{g}$ indicate the velocity error, attitude error, and gyroscope bias error of the Leg-IMU, respectively.

This zero-velocity update is applied to the Leg-IMUs only when foot contact is detected. However, if all IMUs, including the Body-IMU, report zero-velocity, the robot is considered to be in a static state, and the zero-velocity update is performed for all IMUs.

\subsubsection{Relative Position Constraints}
In our proposed DogLegs system, all the Body-IMU and Leg-IMUs perform the strapdown INS propagartion to predict their respective states. However, rather than moving independently, these IMUs are physically constrained within the robot's kinematic chain~\cite{murray2017mathematical}. Their relative positions can be computed in real time using joint encoder readings and known robot parameters. This information helps to constrain the error drift of individual IMUs, improving overall state estimation accuracy. 

Since the multiple IMUs are not aligned together, each has its own independent initial reference frame in which the IMU pose is expressed. As shown in Fig.~\ref{fig:coordinates}, the \textit{world}-frame is defined with the \textit{body}-frame at the initial position. Therefore, to determine the Leg-IMU's pose in the \textit{world}-frame, we must compute the transformation between the initial Leg-IMU frame and the initial Body-IMU frame, namely, \textit{world}-frame.

Taking one Leg-IMU as an example, its transformation w.r.t. the \textit{world}-frame is represented by the rotation matrix $\mathbf{R}^{w}_{\text{limu}}$ and the translation vector $\bm{p}^{w}_{\text{limu}}$. The initial roll and pitch components of $\mathbf{R}^{w}_\text{limu}$ can be determined by IMU data during the initial stationary period of the robot, while the initial translation $\bm{p}^{w}_\text{limu}$ can be obtained from leg kinematic model during the same period. For the initial value of the yaw component of $\mathbf{R}^{w}_\text{limu}$, we assume it to be zero, given that the IMUs are carefully installed and aligned during setup. Therefore, given this transformation and considering the position of the foot contact point in the Leg-IMU frame $\hat{\bm{l}}_{f}^\text{limu}$, the Leg-IMU can estimate the position of the foot in the \textit{world}-frame, denoted as $\hat{\bm{p}}^{w}_f$.

Consequently, the position of the Body-IMU in the \textit{world}-frame, as estimated by the Leg-IMU, can be expressed as
\begin{equation}
    \hat{\bm{p}}^{w}_{b} = \hat{\bm{p}}^{w}_f - \hat{\mathbf{R}}^{w}_{b}\hat{\bm{p}}_{f}^{b},
\end{equation}
where $\hat{\bm{p}}^{w}_{b}$ is the position of the Body-IMU in \textit{world}-frame, and $\hat{\bm{p}}_{f}^{b}$ is the position of the foot in the \textit{body}-frame. By rearranging the equation above, we obtain the relative position of the foot in the \textit{body}-frame, indicated by the Leg-IMU and Body-IMU, as
\begin{equation}
    \hat{\bm{p}}^{b}_{f} = \hat{\mathbf{R}}^{b}_{w}\left(\hat{\bm{p}}^{w}_f - \hat{\bm{p}}^{w}_{b}\right).
\end{equation}
At the same time, for each leg of the robot, the position of the foot in the \textit{body}-frame can be calculated by the leg kinematics model as
\begin{equation}
    \tilde{\bm{p}}^{b}_{f} = \textit{g}(\hat{\bm{\phi}}) = \bm{p}^{b}_{f} - \bm{n}_{p^b_{f}},
\end{equation}
where $\hat{\bm{\phi}}$ is the vector containing all joint angle measurements from the joint encoders of the leg, $\textit{g}(\cdot)$ is the kinematics function of the leg~\cite{murray2017mathematical}, $\bm{p}^{b}_{f}$ indicates the true position of the foot in the \textit{body}-frame, and $\bm{n}_{p^b_{f}}$ indicates the uncertainty of the relative position measurement, which is determined by the noise of the joint encoders and the uncertainty of the leg parameters.
Consequently, we can build the relative position constraint model as 
\begin{equation}
    \delta\bm{z}_p = \hat{\bm{p}}^{b}_{f} - \tilde{\bm{p}}^{b}_{f}.
\end{equation}


Whenever a Leg-IMU detects foot contact, we perform both zero-velocity updates and relative position updates between that Leg-IMU and Body-IMU.

\begin{figure}[t]
    \centering
    \includegraphics[width=8cm]{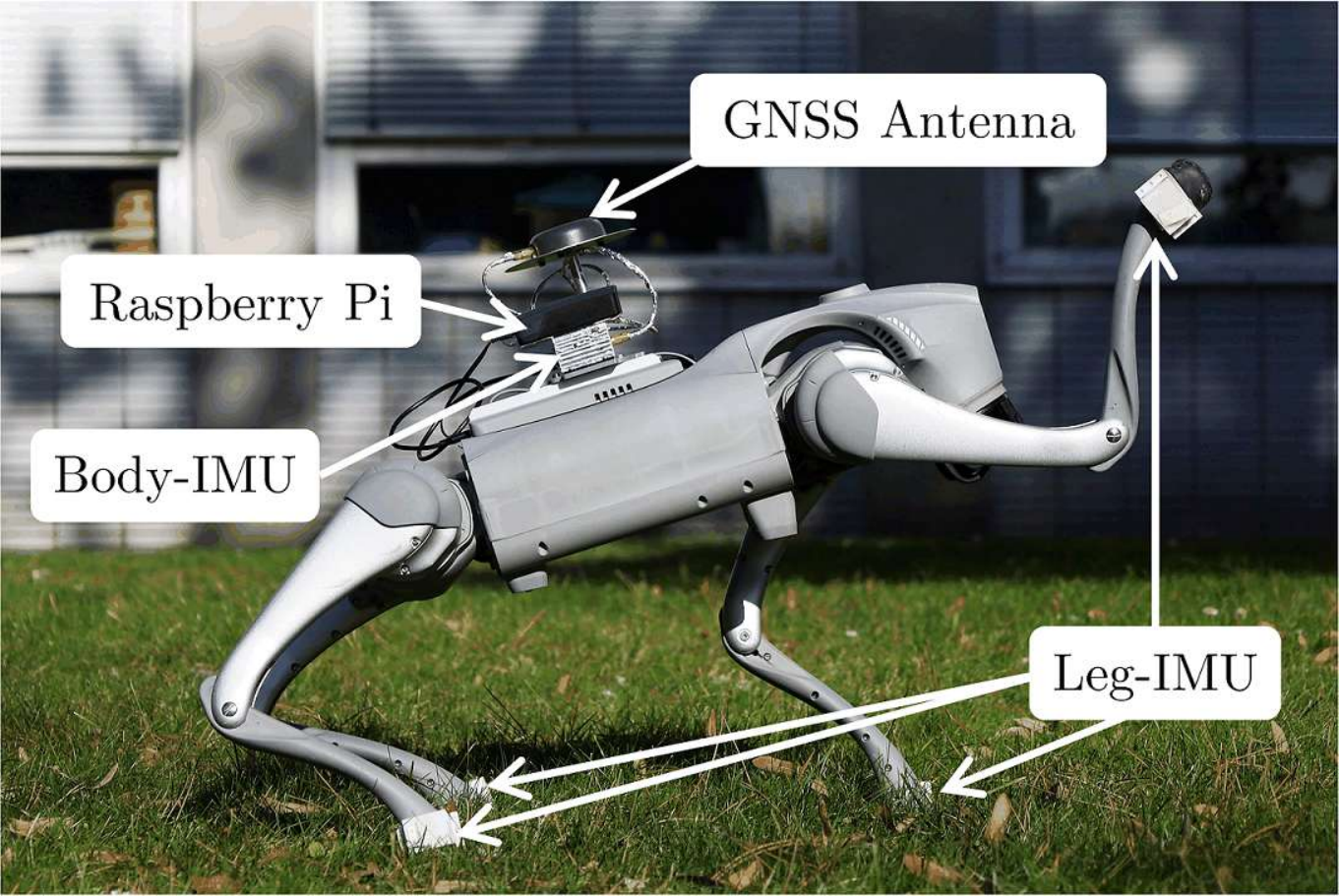}
    \caption{Experimental setup showcasing the Unitree Go2 quadruped robot and the associated devices used in the experiments.}
    \label{fig:experimental_robot}
\end{figure}

\section{Experimental Results}
This section presents real world experimental results to support our key claims, namely, (i) Doglegs achieves better performance in state estimation compared to traditional EKF-based leg odometry across different terrains, and (ii) DogLegs reduces the error drift of the individual IMU systems by incorporating the relative position constraints between the multiple IMUs.

\begin{figure*}[t]
    \centering
    \subfigure[Asphalt road.]{\includegraphics[width=0.19\textwidth]{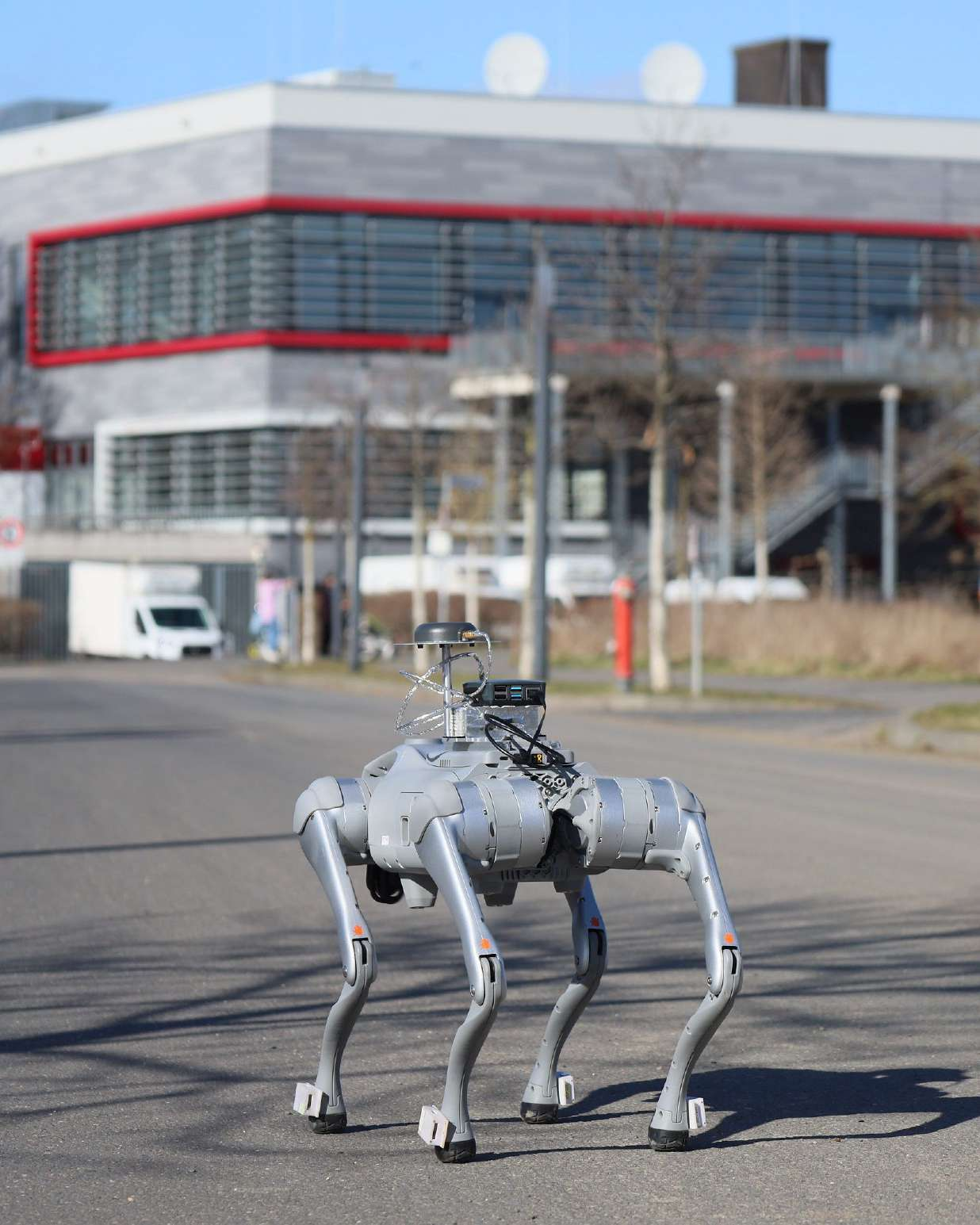}}
    \subfigure[Parking lots.]{\includegraphics[width=0.19\textwidth]{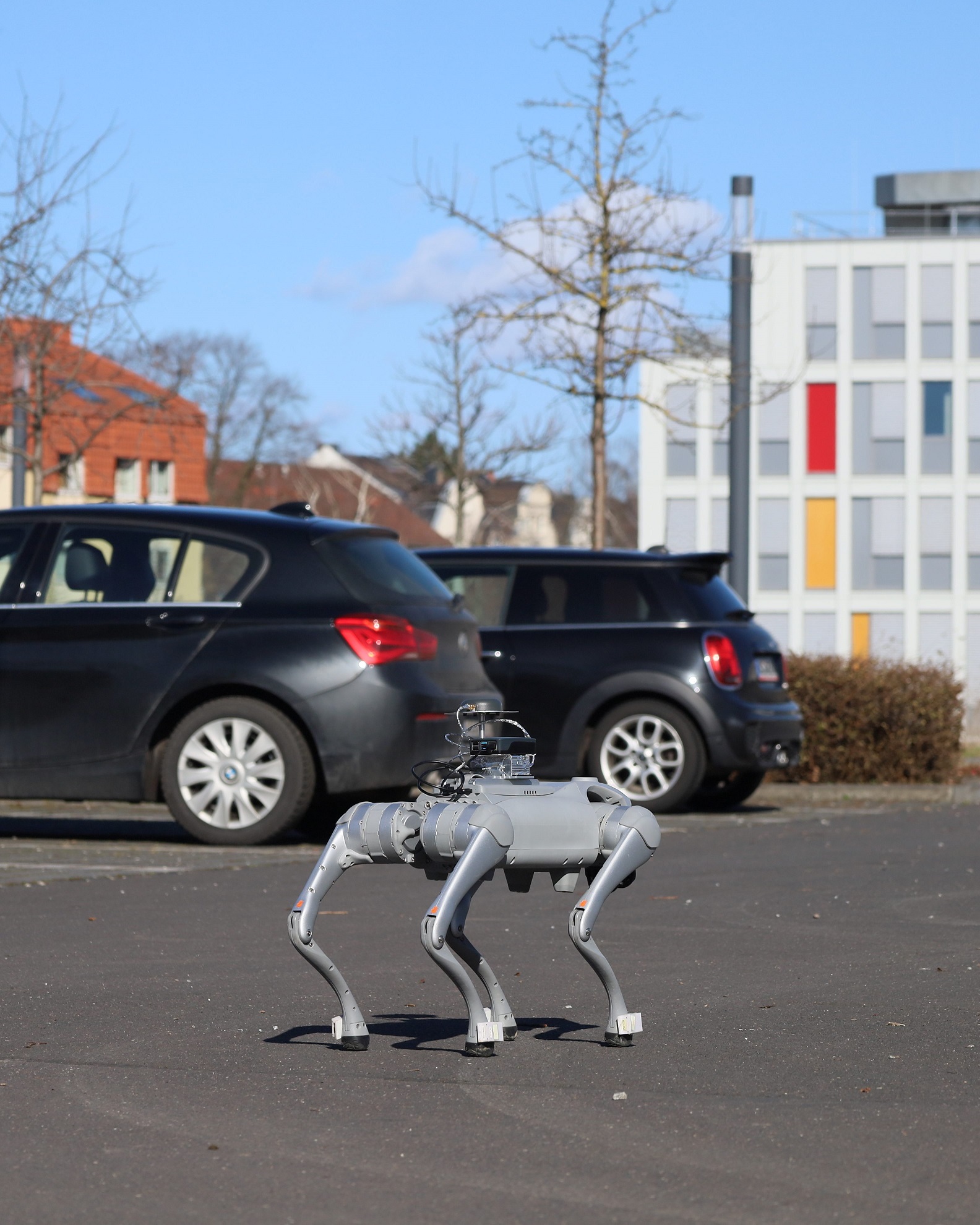}}
    \subfigure[Construction site.]{\includegraphics[width=0.19\textwidth]{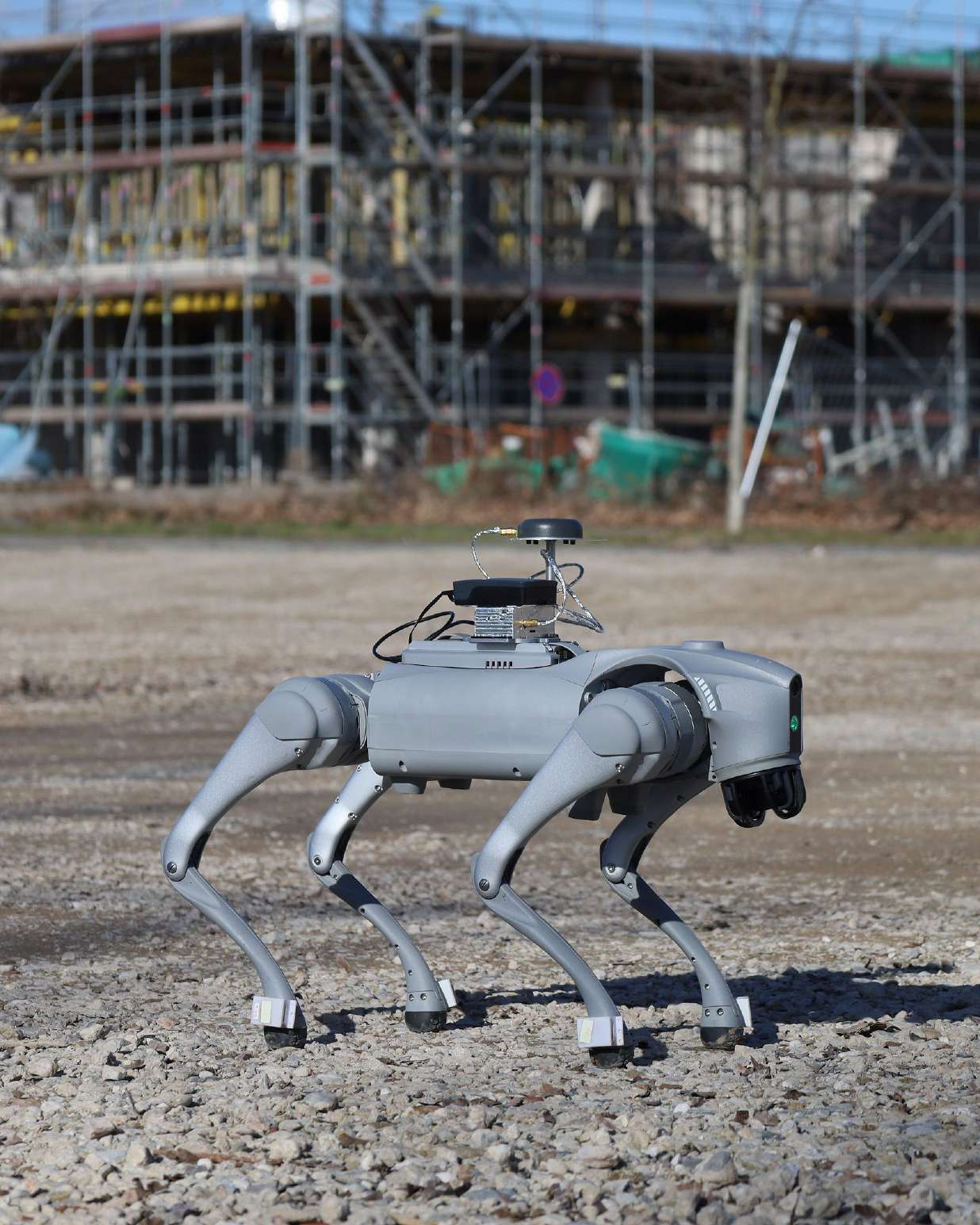}}
    \subfigure[Offroad.]{\includegraphics[width=0.19\textwidth]{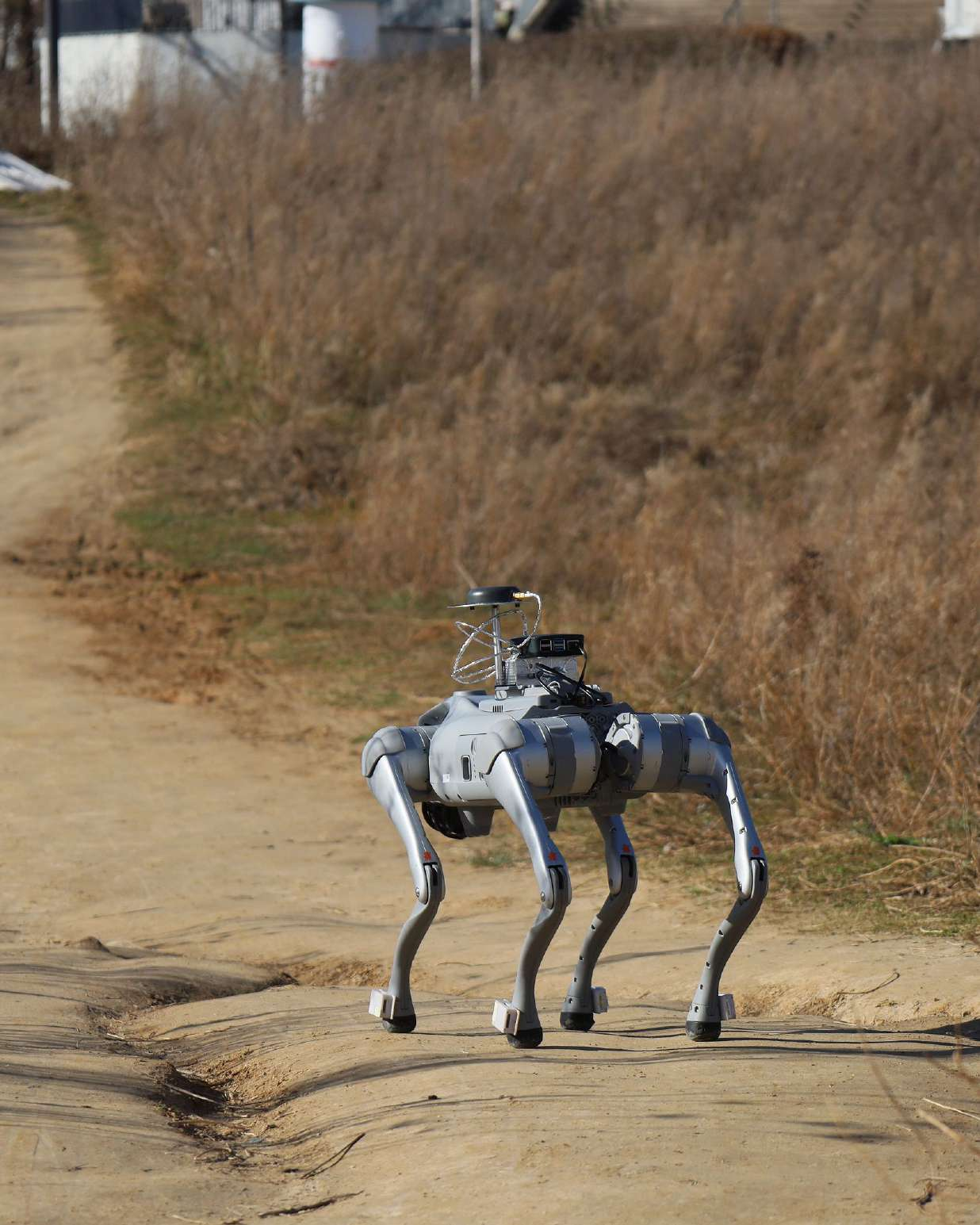}}
    \subfigure[Grass.]{\includegraphics[width=0.19\textwidth]{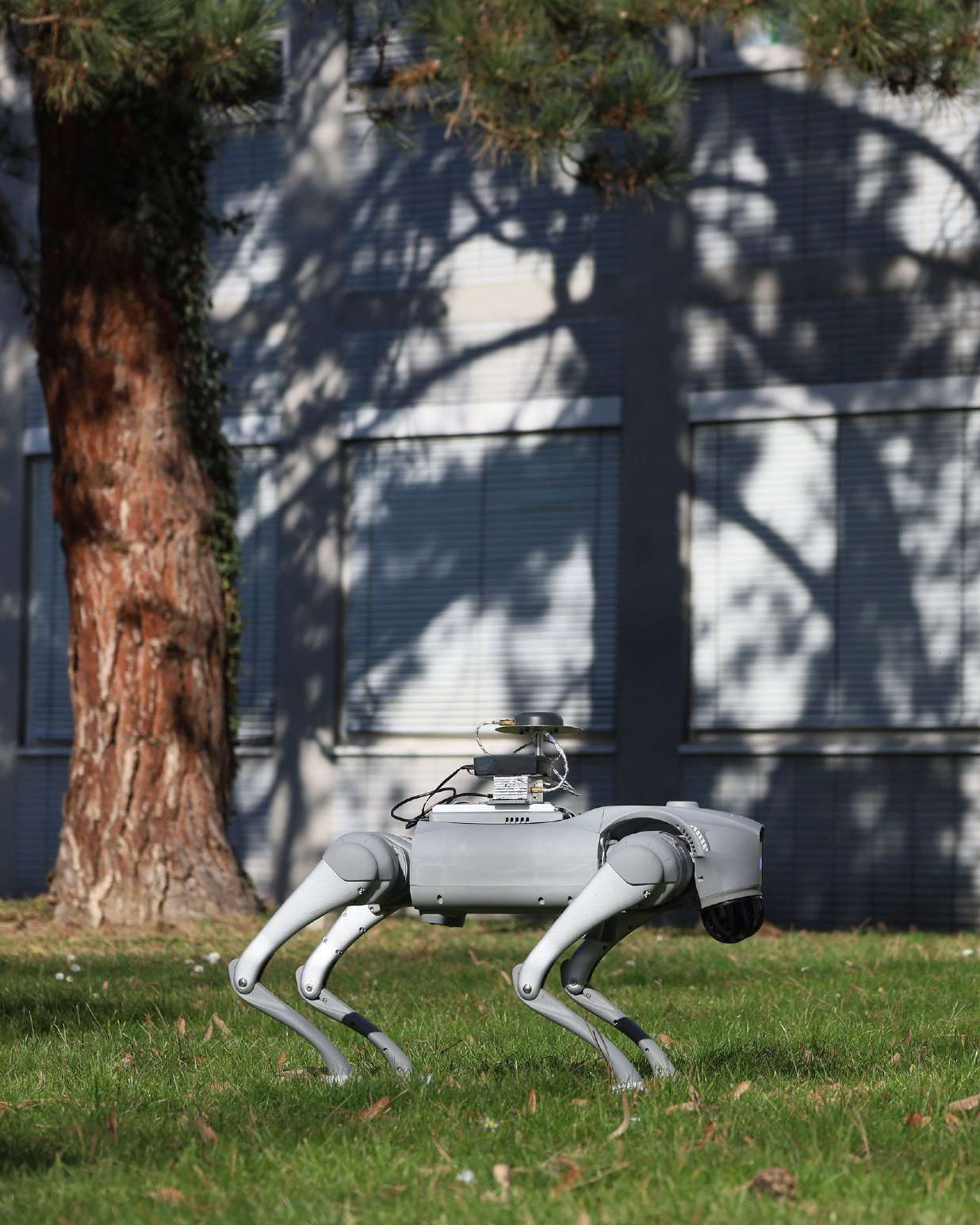}}
    \caption{The different experimental environments of the five sequences.}
    \label{fig:environments}
\end{figure*} 

\subsection{Experimental Setup}
We conducted experiments using a Unitree Go2 quadruped, which was equipped with a Body-IMU, four Leg-IMUs, a Global Navigation Satellite System (GNSS)/IMU integrated navigation system, and a raspberry pi for data collection, as shown in Fig.~\ref{fig:experimental_robot}. The Body-IMU and Leg-IMUs were of the same type, containing a low-cost ICM20602 inertial sensor\footnote{\url{https://invensense.tdk.com/download-pdf/icm-20602-datasheet/}} (approximately 3 US dollars), a rechargeable battery, and a Bluetooth module for data transmission. The Leg-IMUs were mounted on the lower legs of the robot and close to the feet, while the Body-IMU was mounted on the robot body. The extrinsic transformation between the Body-IMU frame and \textit{body}-frame was measured in advance. 

\begin{table}[t]
    \centering
    \begin{threeparttable}
    \caption{Quantitative Comparison on the Pose Accuracy in Four Sequences}
    \label{tab:pose_comparison}
    \begin{tabular}{cccccc}
        \toprule
        \textbf{Sequence} & \textbf{Method} & \textbf{\makecell{RPE\\tra.}} & \textbf{\makecell{RPE\\rot.}} & \textbf{\makecell{APE\\tra.}} & \textbf{\makecell{APE\\rot.}} \\
        \midrule
        \specialrule{0em}{2pt}{2pt}
        \multirow{3}*{\makecell{Asphalt road}} & {Foot-INS} & $1.21$ & $0.31$ & $6.09$ & $2.54$ \\
        & {LegOdom} & $\mathbf{0.70}$ & $0.22$ & ${5.76}$ & $2.12$ \\
        & {DogLegs} & $\mathbf{0.70}$ & $\mathbf{0.17}$ & $\mathbf{4.07}$ & $\mathbf{1.82}$ \\
        \midrule
        \specialrule{0em}{2pt}{2pt}
        \multirow{3}*{\makecell{Parking lots}} & {Foot-INS} & $1.26$ & $0.34$ & $4.46$ & $2.22$ \\
        & {LegOdom} & $\mathbf{0.69}$ & $0.22$ & $\mathbf{2.94}$ & $2.01$ \\
        & {DogLegs} & ${0.76}$ & $\mathbf{0.21}$ & $3.51$ & $\mathbf{1.91}$ \\
        \midrule
        \specialrule{0em}{2pt}{2pt}
        \multirow{3}*{\makecell{Construction site}} & {Foot-INS} & $1.05$ & $0.28$ & $2.29$ & $2.22$ \\
        & {LegOdom} & $0.73$ & $0.20$ & ${2.15}$ & $2.33$ \\
        & {DogLegs} & $\mathbf{0.68}$ & $\mathbf{0.18}$ & $\mathbf{2.01}$ & $\mathbf{2.05}$ \\
        \midrule
        \specialrule{0em}{2pt}{2pt}
        \multirow{3}*{\makecell{Offroad}} & {Foot-INS} & $1.11$ & $0.31$ & $2.72$ & $2.52$ \\
        & {LegOdom} & $\mathbf{0.48}$ & $0.17$ & ${2.47}$ & $1.83$ \\
        & {DogLegs} & ${0.54}$ & $\mathbf{0.12}$ & $\mathbf{2.04}$ & $\mathbf{1.45}$ \\
        \bottomrule
    \end{tabular}   
    \begin{tablenotes}   
        \footnotesize
        \item[*] RPE tra. and RPE rot. denote the relative translation error (\%) and relative rotation error (deg/m), respectively. APE tra. and APE rot. denote the absolute translation error (m) and absolute rotation error (deg), respectively~\cite{grupp2017evo}. The best performance is highlighted in bold.
    \end{tablenotes}
    \end{threeparttable}
\end{table}

\begin{figure}[t]
	\centering
    \begin{minipage}[b]{0.49\linewidth}
        \centering
        \subfigure[Asphalt road.]{
        \includegraphics[width = \linewidth]{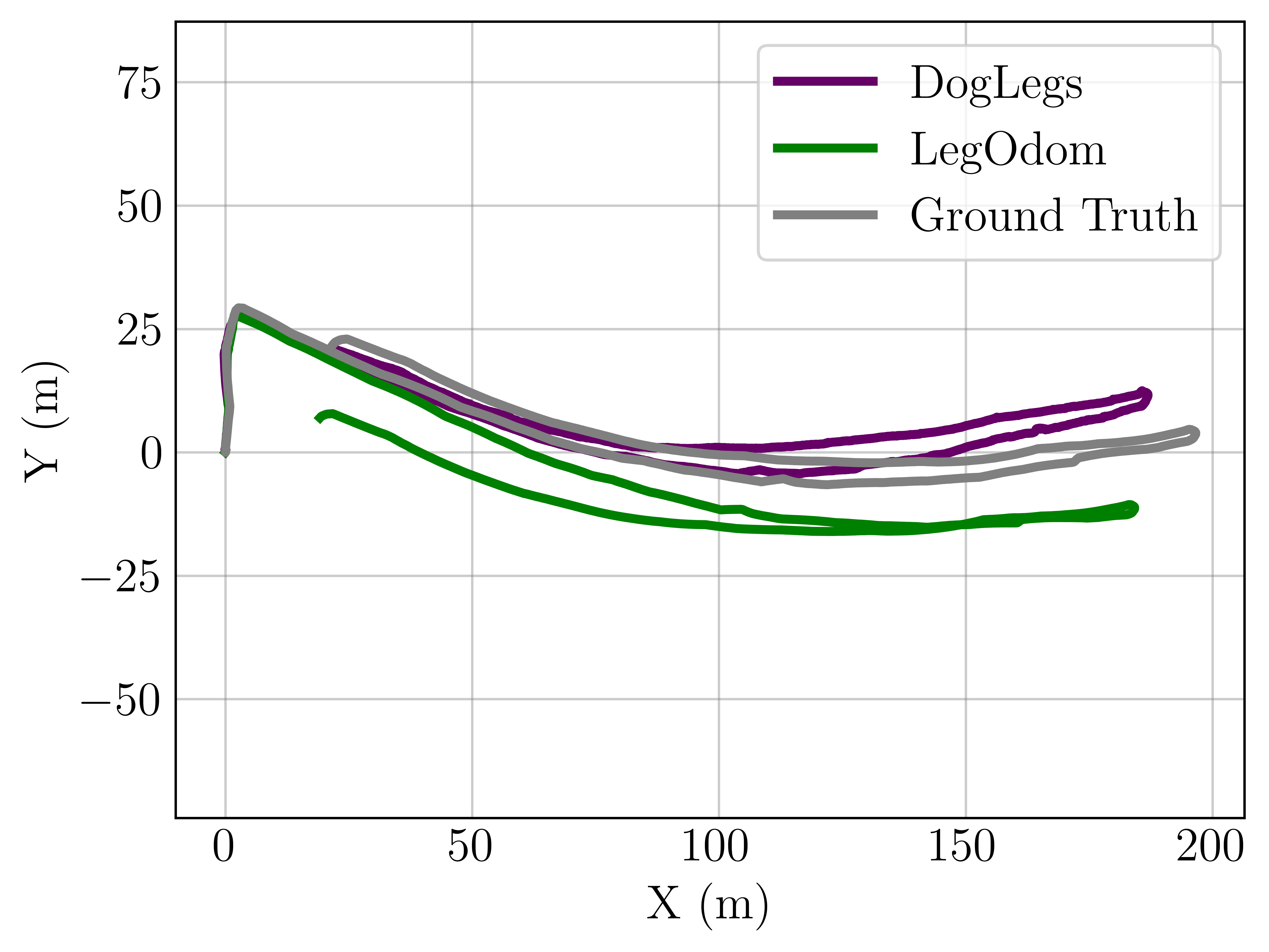}
    }
    \end{minipage}
    \hfill
    \begin{minipage}[b]{0.49\linewidth}
        \centering
        \subfigure[Parking lots.]{
        \includegraphics[width = \linewidth]{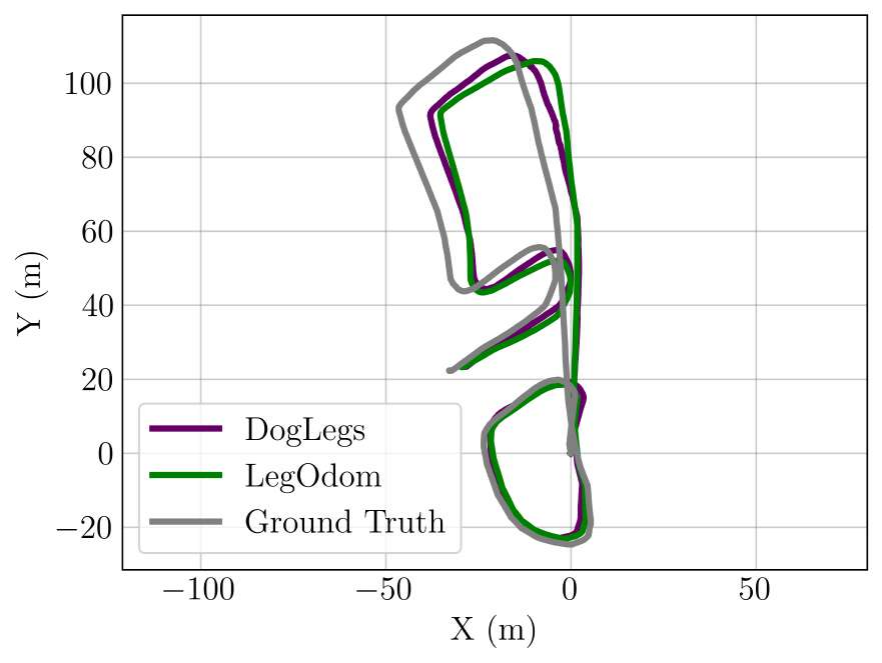}
    }
    \end{minipage}
	
    \begin{minipage}[b]{0.49\linewidth}
        \centering
        \subfigure[Construction site.]{
            \includegraphics[width = \linewidth]{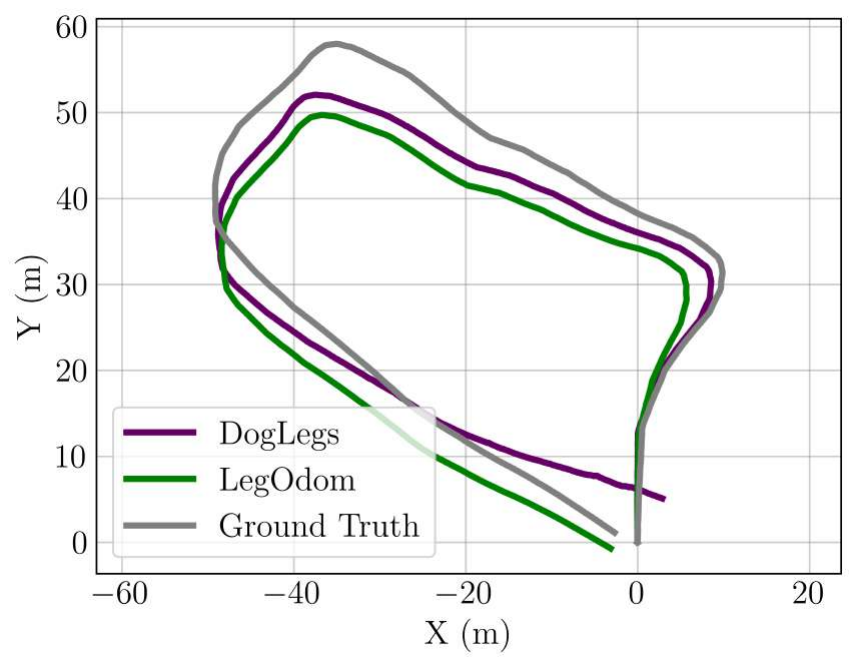}
        }
    \end{minipage}
    \hfill
    \begin{minipage}[b]{0.49\linewidth}
        \centering
        \subfigure[Offroad.]{
            \includegraphics[width = \linewidth]{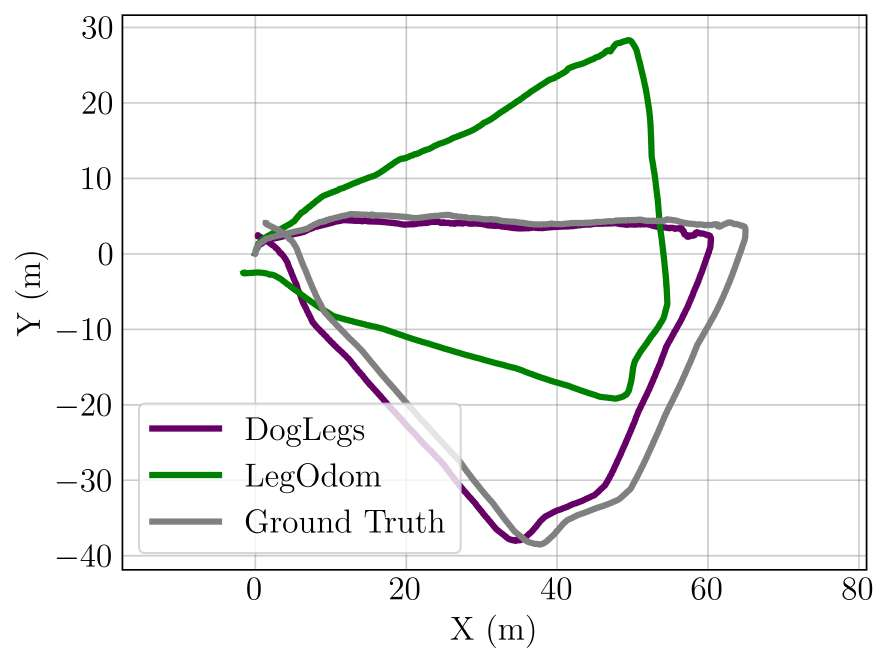}
        }
    \end{minipage}
    \caption{The estimated trajectories of LegOdom and our proposed DogLegs against ground truth in four sequences. As shown, DogLegs demonstrates more accurate trajectory estimation with reduced error drift compared to LegOdom.}
    \label{fig:trajectories}
\end{figure}

We collected datasets in five different outdoor environments, including asphalt road, parking lots, construction site, offroad, and grass. Fig.~\ref{fig:environments} shows the five different environments. The robot was set to walking mode for all experiments. We performed a smoothed GNSS post-processed kinematic (PPK)/IMU integration method to compute the near-ground-truth trajectories for the first four sequences. However, for the grass experiment, GNSS signal were blocked by trees and buildings, rendering them unusable. Instead, we controlled the robot return to its starting point, allowing us to evaluate the loop closure error. The IMU and joint encoder data were collected at \SI{200}{Hz} in all the experiments.

\subsection{State Estimation Performance}

In this section, we analyze the pose estimation accuracy of our method and its generalizability across different terrains. We compare our method with two baselines:
\begin{itemize}
    \item \textbf{LegOdom}. The traditional EKF-based leg odometry\footnote{\url{https://github.com/YibinWu/leg-odometry}} using a Body-IMU and joint encoders~\cite{bloesch2013ekf, bloesch2013ukf}.
    \item \textbf{Foot-INS}. Single Foot-IMU based localization system~\cite{niu2022iot}. 
\end{itemize}

We do not include a comparison with Yang~\etalcite{yangshuo2023mipo} because it drifted quickly and failed to produce meaningful results in our experiments. To ensure a fair comparison, we assign the roll and pitch angles estimated by DogLegs to Foot-INS, as the Leg-IMUs exhibit different roll and pitch orientations relative to the main body. In addition, we report the mean value of the four individual systems on the four legs for Foot-INS. Table~\ref{tab:pose_comparison} lists the relative pose error (RPE) and absolute pose error (APE) calculated using evo~\cite{grupp2017evo} in four sequences. All the estimated trajectories were aligned with the ground truth using the transformation computed by evo~\cite{grupp2017evo} before metric evaluation. 

\begin{figure}[t]
    \centering
    \includegraphics[width=8.8cm]{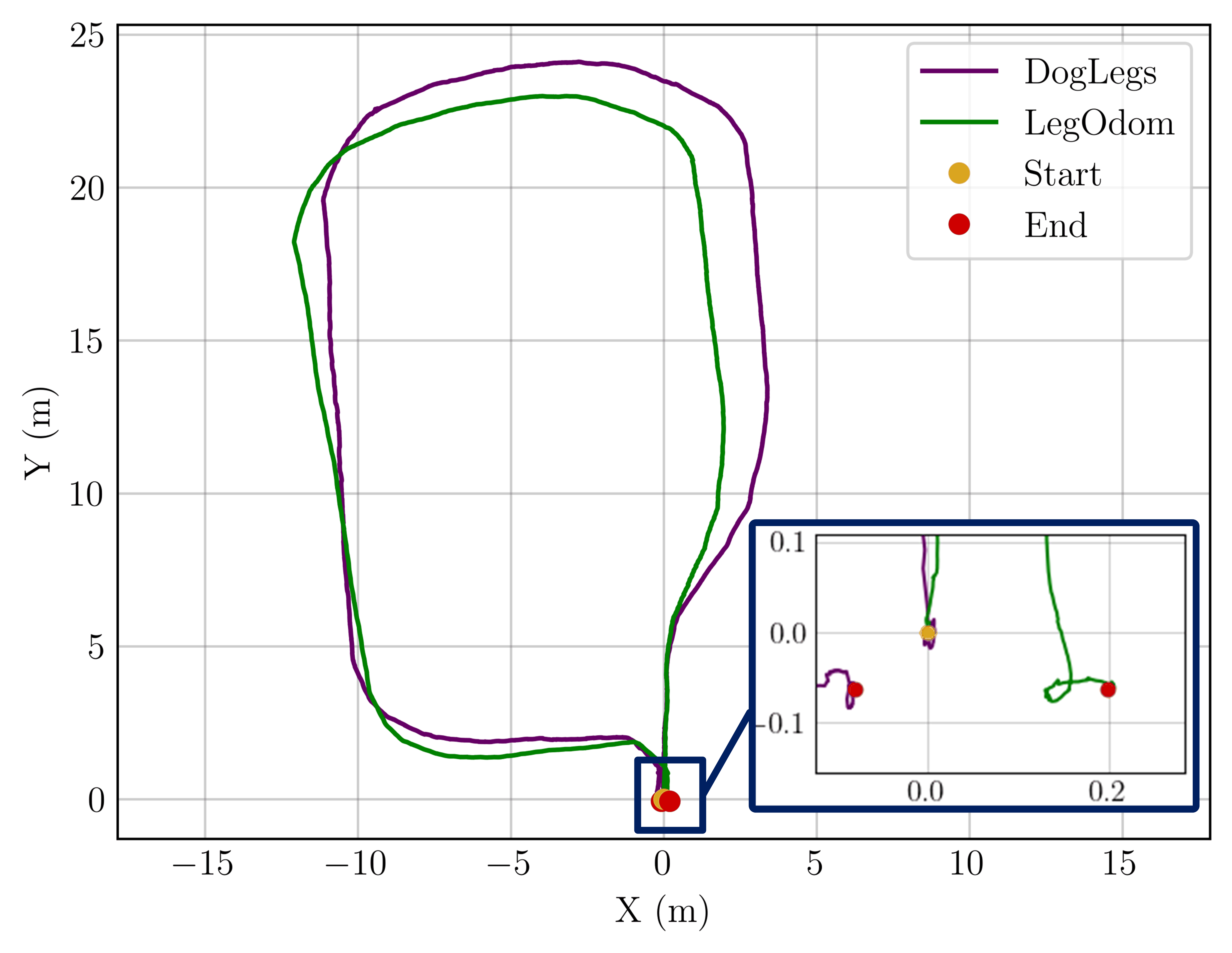}
    \vspace{-0.45em}
    \caption{The estimated trajectories of LegOdom and our proposed DogLegs in the grass experiment. Because this area is surrounded by buildings and trees, GNSS could not provide useful positioning results. We evaluate the loop closure error in this experiment. As can be seen, DogLegs shows a smaller error than LegOdom when the robot returns to its starting point.}
    \label{fig:grass_trajectories}
\end{figure}

\begin{table}[t]
    \centering
    \begin{threeparttable}
    \caption{Quantitative Comparison on the Roll and Pitch Accuracy in Four Sequences}
    \label{tab:roll_pitch_comparison}
    \begin{tabular}{cccccc}
        \toprule
        \multirow[c]{2}*{\makecell{\textbf{Sequence}}} & \multirow[c]{2}*{\makecell{\textbf{Method}}} & \multicolumn{2}{c}{\textbf{Roll} (\SI{}{\degree})} & \multicolumn{2}{c}{\textbf{Pitch} (\SI{}{\degree})} \\
        \cmidrule(lr){3-6}
        &  & \textbf{RMSE} & \textbf{MAX} & \textbf{RMSE} & \textbf{MAX} \\
        \midrule
        \specialrule{0em}{2pt}{2pt}
        \multirow{2}*{\makecell{Asphalt road}} 
        & {LegOdom} & $2.79$ & $5.95$ & $\mathbf{0.67}$ & $5.41$ \\
        & {DogLegs} & $\mathbf{0.59}$ & $\mathbf{1.46}$ & ${0.75}$ & $\mathbf{1.36}$ \\
        \midrule
        \specialrule{0em}{2pt}{2pt}
        \multirow{2}*{\makecell{Parking lots}} 
        & {LegOdom} & ${1.52}$ & $3.50$ & ${1.03}$ & $3.50$ \\
        & {DogLegs} & $\mathbf{0.81}$ & $\mathbf{1.71}$ & $\mathbf{0.57}$ & $\mathbf{1.47}$ \\
        \midrule
        \specialrule{0em}{2pt}{2pt}
        \multirow{2}*{\makecell{Construction site}} 
        & {LegOdom} & $0.69$ & $1.71$ & ${0.92}$ & $2.01$ \\
        & {DogLegs} & $\mathbf{0.50}$ & $\mathbf{1.19}$ & $\mathbf{0.74}$ & $\mathbf{1.45}$ \\
        \midrule
        \specialrule{0em}{2pt}{2pt}
        \multirow{2}*{\makecell{Offroad}} 
        & {LegOdom} & ${0.90}$ & $2.00$ & $\mathbf{0.63}$ & $1.80$ \\
        & {DogLegs} & $\mathbf{0.49}$ & $\mathbf{1.17}$ & ${0.85}$ & $\mathbf{1.43}$ \\
        \bottomrule
    \end{tabular}   
    \begin{tablenotes}   
        \footnotesize
        \item[*] RMSE and MAX denote the root mean square error and max error, respectively. The best performance is highlighted in bold.
    \end{tablenotes}
    \end{threeparttable}
\end{table}

\begin{figure}[t]
    \centering
    \includegraphics[width=8.8cm]{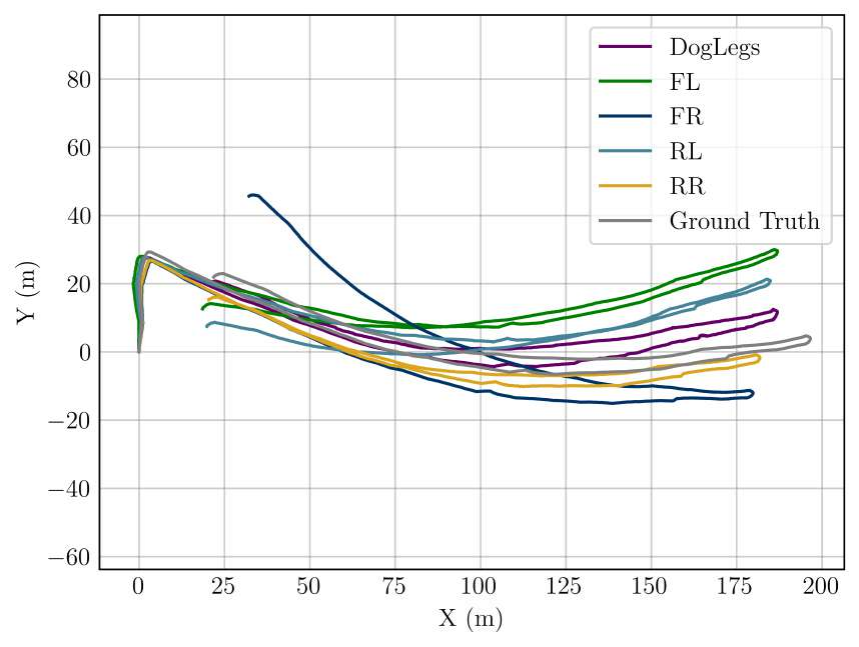}
    \caption{The estimated trajectories of the individual Leg-IMU systems (FL, FR, RL, RR) and our proposed DogLegs compared with ground truth in the asphalt road sequence.}
    \label{fig:multiimu}
\end{figure}

Fig.~\ref{fig:trajectories} compares the estimated trajectories of LegOdom and our proposed DogLegs against the ground truth across the four sequences. To illustrate the actual absolute error drift, in this figure, we didn't compute the transformation to align the whole estimated trajectories with the ground truth as done in evo~\cite{grupp2017evo}. 

As shown in Table~\ref{tab:pose_comparison}, DogLegs achieves overall superior performance across all four sequences. However, in the parking lots sequence, it shows a slightly larger translation error than LegOdom, which is likely due to the stochastic error of the inertial sensors. Furthermore, the alignment of the estimated trajectory with the ground truth may partially compensate for actual drift, thereby obscuring the true performance advantage of the proposed DogLegs. This is illustrated in Fig.~\ref{fig:trajectories}~(b), where the trajectory of DogLegs more closely matches the ground truth compared to LegOdom.

In addition, Fig.~\ref{fig:trajectories}~(a) and~(d) clearly show that DogLegs exhibits significantly smaller drift compared to LegOdom. Particularly, in the offroad sequence shown in Fig.~\ref{fig:trajectories}~(d), LegOdom exhibits a large heading error at the beginning of the trajectory, leading to a large drift in the estimated trajectory. In contrast, DogLegs maintains a more stable heading estimation, resulting in a more accurate trajectory. 

In the grass experiment, the GNSS signal was obstructed by trees and buildings, preventing the GNSS/IMU integrated navigation system from providing an accurate trajectory reference. Therefore, we evaluated the robot’s loop closure error when it returned to the starting point. The results, shown in Fig.~\ref{fig:grass_trajectories}, indicate that DogLegs achieves a loop closure distance error of approximately \SI{0.1}{m}, which is smaller than half of LegOdom's error.

In addition to position and yaw angle, accurate estimation of roll and pitch angles is crucial for legged robots to maintain balance and execute dynamic movements effectively. Table~\ref{tab:roll_pitch_comparison} presents the statistic results of the roll and pitch angle errors for LegOdom and DogLegs across the four sequences. We report both the root mean square error (RMSE) and maximum absolute error (MAX). From the table, we can see that overall, DogLegs exhibits lower roll and pitch estimation error than LegOdom. Moreover, DogLegs provides more stable roll and pitch estimates, as indicated by its smaller MAX error. For instance, in both the asphalt road and parking lots sequences, the maximum roll and pitch errors of LegOdom are around three times larger than those of DogLegs. 

In conclusion, these experimental results support our first claim that Doglegs achieves better performance in state estimation compared to traditional EKF-based leg odometry across different terrains.


\subsection{Multi-IMU Constraint}
In this section, we evaluate the effectiveness of our proposed relative position constraint between multiple IMUs. Specifically, we compare the state estimation accuracy of DogLegs against individual IMU systems in the same sequence. This analysis supports our second claim: DogLegs reduces the error drift of the individual IMU systems by incorporating the relative position constraints between the multiple IMUs.

Fig.~\ref{fig:multiimu} compares the trajectories of the individual Leg-IMU systems estimated by Foot-INS~\cite{niu2022iot}, namely, front-left foot (FL), front-right foot (FR), rear left foot (RL), rear right foot (RR) alongside DogLegs and the ground truth in the asphalt road experiment. The results show that the DogLegs trajectory falls approximately in the middle of the multiple independent Leg-IMU trajectories. Due to the random sensor errors inherent in IMUs, the individual Leg-IMU systems exhibit varying levels of error drift in different directions, even using the same type of sensor. By integrating diverse motion information from multiple IMUs, DogLegs effectively reduces random errors from individual IMU systems, thereby providing more accurate and stable state estimates overall.

\begin{figure}[t]
    \centering
    \includegraphics[width=8.8cm]{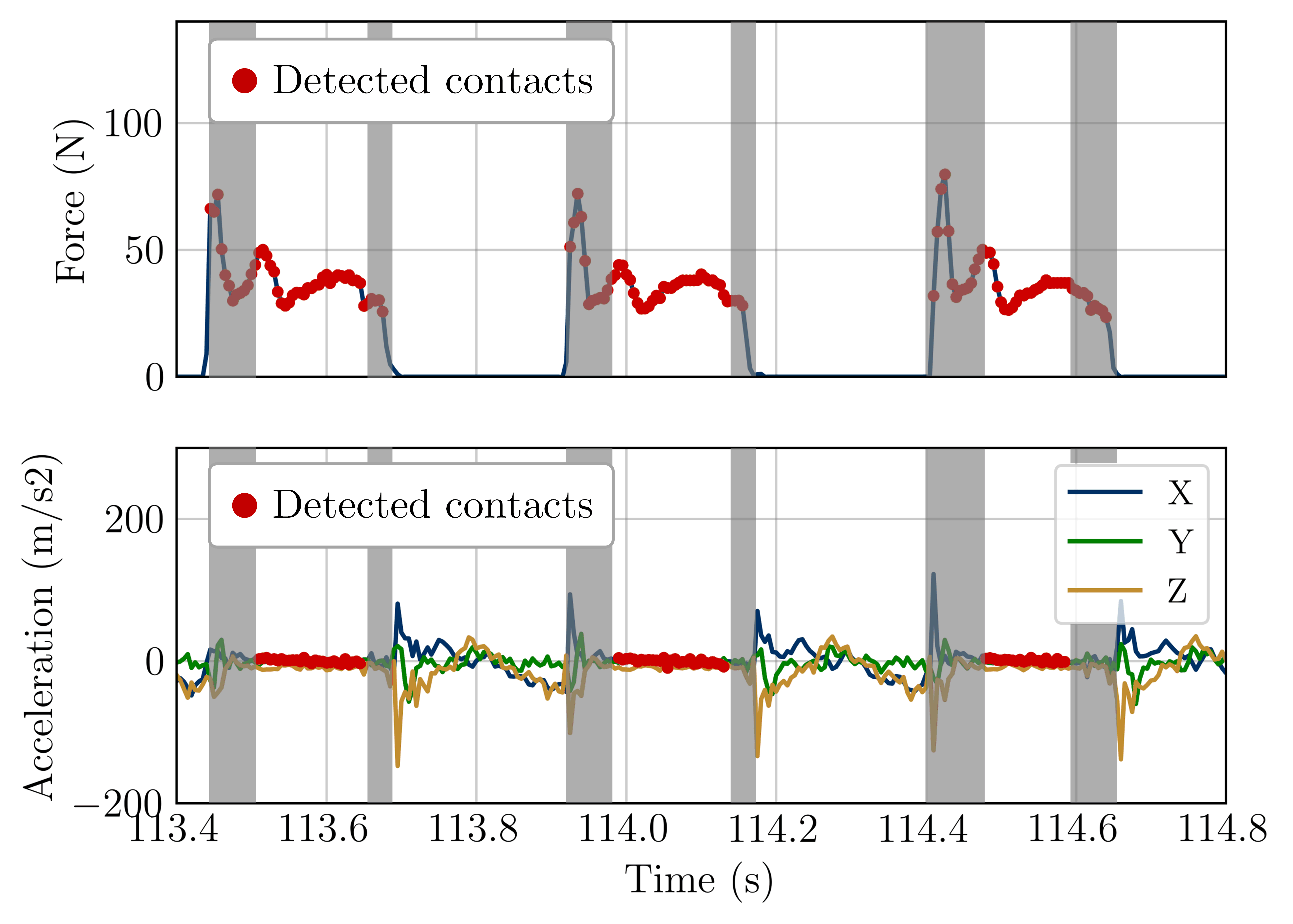}
    \caption{Detection results of front-left foot contact moments with the force threshold-based method and our proposed DogLegs in the asphalt road sequence. The upper and lower subfigures show the detected results using a force sensor (with a threshold set to $20$) and Leg-IMU accelerometer measurements, respectively. The gray areas indicate false positives (upper) and true negatives intervals (lower) in contact detection. As shown, our proposed DogLegs improves the robustness of contact detection by effectively rejecting outliers.}
    \label{fig:zupt_detection}
\end{figure}

\subsection{Foot Contact Detection}
To evaluate the foot contact detection performance of our proposed DogLegs system, we compare its front-left foot contact detection results with those of the force threshold-based method in the asphalt road sequence. As shown in Fig.~\ref{fig:zupt_detection}, DogLegs detects more stable foot contact intervals, while force threshold-based method introduces outliers due to its reliance on a single-dimensional force measurement.

\subsection{Computation Efficiency}
In this section, we analyze the computational efficiency of our proposed DogLegs. All experiments were conducted on a desktop computer equipped with an Intel i7-10700 CPU running at \SI{2.90}{GHz} and \SI{32}{GB} of RAM, using a single computation thread. The average computation time per IMU measurement of DogLegs across all sequences is summarized in Table~\ref{tab:computation_time}. The prediction time includes the strapdown INS propagation for all IMUs as well as the error-state propagation in the EKF. The update time accounts for the processing of all zero-velocity updates and relative position constraints. Note that multiple foot contact events may occur simultaneously. The total time represents the overall time required to process all the IMU measurements and joint encoder received coming at the same time. All values are reported in milliseconds (ms) in Table~\ref{tab:computation_time}.

As shown in Table~\ref{tab:computation_time}, the average computation time of DogLegs is approximately \SI{0.39}{ms} per measurement, which is significantly lower than the IMU measurement interval of \SI{5}{ms} at \SI{200}{Hz}. This indicates that DogLegs can run in real-time on a standard desktop computer. The prediction step takes about \SI{0.19}{ms}, while the update step takes about \SI{0.87}{ms}. The longer duration of the update step is due to the additional computations involved in processing zero-velocity updates and relative position constraints triggered by all foot contact events. Notably, the average total processing time is shorter than the average update time because the update step is not executed at every time step, namely, it is only performed when a foot contact is detected.

\begin{table}[t]
    \centering
    \begin{threeparttable}
    \caption{Average Computation Time per IMU Measurement of DogLegs (ms)}
    \label{tab:computation_time}
    \begin{tabular}{>{\centering\arraybackslash}p{1.6cm} >{\centering\arraybackslash}p{1.6cm} >{\centering\arraybackslash}p{1.6cm}}
        \toprule
        \textbf{Prediction} & \textbf{Update} & \textbf{Total}\\
        \midrule
        $0.19$ & $0.87$ & $0.39$\\
        \bottomrule
    \end{tabular}   
    \end{threeparttable}
\end{table}

\section{Discussion and Conclusion}
Our goal in this study was to develop a robust proprioceptive state estimator for legged robots while minimizing additional hardware and computational costs. To this end, we proposed DogLegs, a framework that fuses the measurements from a Body-IMU, joint encoders, and multiple Leg-IMUs via an error-state EKF. Real world experiments have shown that DogLegs achieves better performance in state estimation compared to the traditional EKF-based leg odometry in different challenging terrains. In addition, DogLegs effectively reduces the error drift of the individual IMU systems by incorporating the relative position constraints between the multiple IMUs.

We emphasize that the IMU chips used in our system are commercial-grade components, costing approximately 3 US dollars only. Although in our current experiments, the IMU are attached to the exterior of the quadruped robot's lower legs, they can be seamlessly integrated into the internal structure without modifying the existing design. Additionally, power supply and data transmission are not limiting factors, as the lower legs already accommodate force sensors. Therefore, our approach can be directly applied to existing legged robots.

Future work includes leveraging Leg-IMUs for foot slippage detection and incorporating foot contact uncertainty into the state estimation framework.

\section{Acknowledgment}
We thank Dr. Shuo Yang, Dr. Tao Liu, and Dazhou Xia for the fruitful discussion on the algorithm. We also appreciate Sai Anudeep Sajja, Delong Tao and Manuel Mittelstedt for their help in conducting the experiments, and Gereon Tombrink for reviewing the paper.

\bibliographystyle{IEEE}
\bibliography{bibliography/IEEEabrv, bibliography/my_abrv, bibliography/doglegs}
\end{document}